\pdfoutput=1
\documentclass[a4paper,11pt]{article}
\usepackage[utf8]{inputenc}
\usepackage[font=small,skip=1pt]{caption}
\usepackage[rmargin = 2.5cm, lmargin = 2.5cm, tmargin = 2.5cm, bmargin = 2.5cm]{geometry}
\usepackage{authblk}
\usepackage{subcaption}
\usepackage{titlesec}
\titleformat{\section}%
  [hang]
  {\normalfont\bfseries\Large}
  {}
  {0pt}
  {}
\usepackage{graphicx}
\usepackage{amsmath,amssymb}
\usepackage{hyperref} 
\hypersetup{
    colorlinks=true,
    linkcolor=black,
    citecolor=black,
    filecolor=black,
    urlcolor=blue,
}
\usepackage{setspace} 

\def    \be            {\begin{equation}}
\def    \ee            {\end{equation}}
\def    \bea           {\begin{eqnarray}}
\def    \eea           {\end{eqnarray}}

\DeclareMathOperator*{\argmax}{arg\,max}

\title{{\bf Deep imagination is a close to optimal policy for planning in large decision trees under limited resources}\vspace{10pt}} 

\author[*1,2]{Rubén Moreno-Bote}
\author[1]{Chiara Mastrogiuseppe}

\affil[1]{Center for Brain and Cognition, and Department of Information and Communication Technologies, Universitat Pompeu Fabra, Barcelona, Spain}
\affil[2]{Serra Húnter Fellow Programme, Universitat Pompeu Fabra, Barcelona, Spain}
\affil[*]{To whom correspondence may be addressed. Email: ruben.moreno@upf.edu}
\date{\today}

\begin{document}
   	
\maketitle

\section*{Abstract}
Many decisions involve choosing an uncertain course of actions in deep and wide decision trees, as when we plan to visit an exotic country for vacation. In these cases, exhaustive search for the best sequence of actions is not tractable due to the large number of possibilities and limited time or computational resources available to make the decision. Therefore, planning agents need to balance breadth (exploring many actions at each level of the tree) and depth (exploring many levels in the tree) to allocate optimally their finite search capacity. We provide efficient analytical solutions and numerical analysis to the problem of allocating finite sampling capacity in one shot to large decision trees.
We find that in general the optimal policy is to allocate few samples per level so that deep levels can be reached, thus favoring depth over breadth search.
In contrast, in poor environments and at low capacity, it is best to broadly sample branches at the cost of not sampling deeply, although this policy is marginally better than deep allocations.
Our results provide a theoretical foundation for the optimality of deep imagination for planning and show that it is a generally valid heuristic that could have evolved from the finite constraints of cognitive systems.

\section*{Introduction}

When we plan our next vacation to an exotic paradise, we decide on a course of actions that has a tree structure: first, choose a country to visit, then the city to stay in, what restaurant or show to go, and so on. 
Planning is a daunting problem because the number of scenarios that could be considered grows exponentially with the depth and width of the associated decision tree.
The dilemma that arises then is how to allocate limited search resources over large decision trees: should we consider many countries for our next vacation (breadth), at the cost of not evaluating very thoroughly any of them, or should we consider very few countries more deeply (depth), at the risk of missing the most exciting one? The above problem is one example of the so-called \emph{breadth-depth} (BD) dilemma, important in tree search algorithms \cite{horowitz_fundamentals_1978,korf_depth-first_1985}, optimizing menu designs \cite{miller_depth/breadth_1981}, decision-making \cite{moreno-bote_heuristics_2020,ramirez-ruiz_optimal_2021,vidal21}, knowledge management \cite{turner_exploring_2002} and education \cite{schwartz_depth_2009}.

Optimizing BD tradeoffs in decision trees is a hard problem due to the combinatorial explosion of states with their depth (number of levels) and width (number of actions per node).
Many approaches that work in relatively small trees, do not scale well in large decision trees where BD tradeoffs will be most relevant. For instance,   
optimal policies in decisions trees can be found by solving the Bellman equation using backwards induction \cite{sutton_reinforcement_1998}, but exact induction is intractable in very large trees due to the exponential grow of states with depth of the tree. Monte Carlo tree search algorithms \cite{browne_survey_2012} approximate the optimal solution by efficiently exploring promising tree nodes, but these methods assign in the long run a non-zero sampling probability to every available action, and thus they do not scale well in very wide decision trees. Meta-reasoning approaches extend the notion of actions to any internal action that can update the state of knowledge of the agent, such as expanding a node in a decision tree and simulating its value \cite{russell_principles_1991,gershman_computational_2015,griffiths_rational_2015,callaway_human_2021,sezener_optimizing_2019,keramati_adaptive_2016}, but as they are formally identical to dynamic programming \cite{drugowitsch_cost_2012,ortega_information-theoretic_2015}, exact inference is extremely expensive in large trees. 

While the above approaches will sample exhaustively all tree nodes in the long run, exhaustive search might be prohibitive, unnecessary or both. First, agents are characterized by having finite capacity \cite{russell_principles_1991,gershman_computational_2015,griffiths_rational_2015,moreno-bote_heuristics_2020, patel2020dynamic, malloy2020deep}, and thus in practice any algorithm needs to be aware of the limited resources available. Second, not every action in a node needs to be sampled in order to achieve a relatively high performance, and thus in practice many actions might be ignored from the very outset of the planning process.
For example, in economic choices the first relevant decision is to select a small consideration set out of the many options available and then make a choice within the smaller set, a heuristic that pervades human behavior 
\cite{hauser_evaluation_1990-1,stigler_economics_1961,roberts_development_1991-1,mehta_price_2003-1,santos_testing_2012-1,scheibehenne_can_2010-1}. Many other situations are best characterized by the availability of compound actions where a myriad of simple actions can be performed in parallel with little or no interaction between them. Examples range from cognitive systems where millions of neurons can perform independent computations in parallel, over investing, where money can be divided and allocated in a combinatorial number of ways, to social decisions \cite{pratt_quorum_2002}. In all these cases, exhaustive exploration of all possible actions and levels is prohibitive. 

Optimization of BD tradeoffs have been studied using the framework of infinitely many-armed bandits and combinatorial multi-armed bandits where finite resources can be arbitrarily allocated among many options. These include one-shot infinitely many-armed Bernoulli \cite{moreno-bote_heuristics_2020} and Gaussian \cite{ramirez-ruiz_optimal_2021} bandits with compound actions, sequential infinitely many-armed Bernoulli  bandits \cite{berry93} and broader families thereof \cite{wang08} with simple actions, and sequential combinatorial multi-armed bandits with compound actions \cite{chen_combinatorial_2018}. These studies show that it is indeed optimal to ignore the vast majority of options while focusing sampling on a relatively small number of them that sublinearly scales with capacity \cite{moreno-bote_heuristics_2020,ramirez-ruiz_optimal_2021}. However, the described optimal BD tradeoffs have been limited to trees of depth one, and thus how to balance breadth and depth search in decision trees remains an unresolved problem. 

In this paper we characterize the optimal sampling policies in model-based planning for the allocation of finite search capacity over a large, stochastically and binarily rewarded, decision tree (Fig. 1). 
Rewards resulting from visiting the tree nodes are unknown and can be learned by sampling them, but as a finite, possibly low, number of samples are available, the agent needs to determine the best way to allocate them over the nodes of the tree. The agent, if desired, could allocate many samples in the first levels, but then search capacity will be exhausted without reaching deep into the tree (breadth search; Fig. 1a), or could allocate few samples per level such that the tree can be sampled deeply (depth search; Fig. 1a), or anything in between. 
We consider the problem of allocating samples in one shot without knowing their individual outcomes, corresponding to a single compound action that consists of many simple actions to be executed in parallel. 
One-shot allocations describe situations where the dispatching of sampling resources needs to be made before feedback is received. Even if the assumption of long delays does not hold, our framework will still be relevant when it is better to use simpler allocation strategies that are agnostic to feedback and thus avoid computational overload. 
In the trip example, a one-shot allocation policy would correspond to sampling countries, cities, etc. in magazines or books, independently of each other during some period of time, and having decided beforehand how many countries, cities, etc. will be sampled. 
Once all the information is acquired, the agent could choose the best course of actions. 
Thus, optimal allocations are sampling policies that maximize the probability of finding the best course of actions starting at the root of the tree by using only the information obtained from the samples. 

We describe the optimal sampling policy over large decision trees as a function of the capacity of the agent and the difficulty of obtaining rewards. We develop an efficient \emph{diffusion-maximization} algorithm for the exact evaluation of the search policies with computational cost of order $\mathcal{O}(bd^2)$, where $d$ is the number of levels of the decision tree and $b$ is its branching factor, much better than the scaling $\mathcal{O}(b^d)$ using backwards induction on the tree itself.
We find that it is generally better to sample very deeply the decision tree such that information over many levels can be gathered, a policy that we call {\em deep imagination}, in analogy to human imagination.   
We find that the optimal number of actions that are explored per node is just two in most conditions, thus leading to a vast options-narrowing effect by which most available actions per node are ignored from the outset of the planning process.  
Regardless of capacity, in rich environments it is best to allocate samples deeply into many levels, such that depth is favored over breadth, and departures from the optimal policy result in large performance impairments. In poor environments at low capacity, it is best to broadly sample branches at the cost of not sampling deeply, although this policy is very often only marginally better than deep allocations. All together, our results provide a theoretical foundation for the optimality of deep imagination for model-based planning in large decision trees, which will be discussed in relation to similar heuristics used in human planning.

\section*{Results}

\begin{figure}
\includegraphics[width=0.7\textwidth]{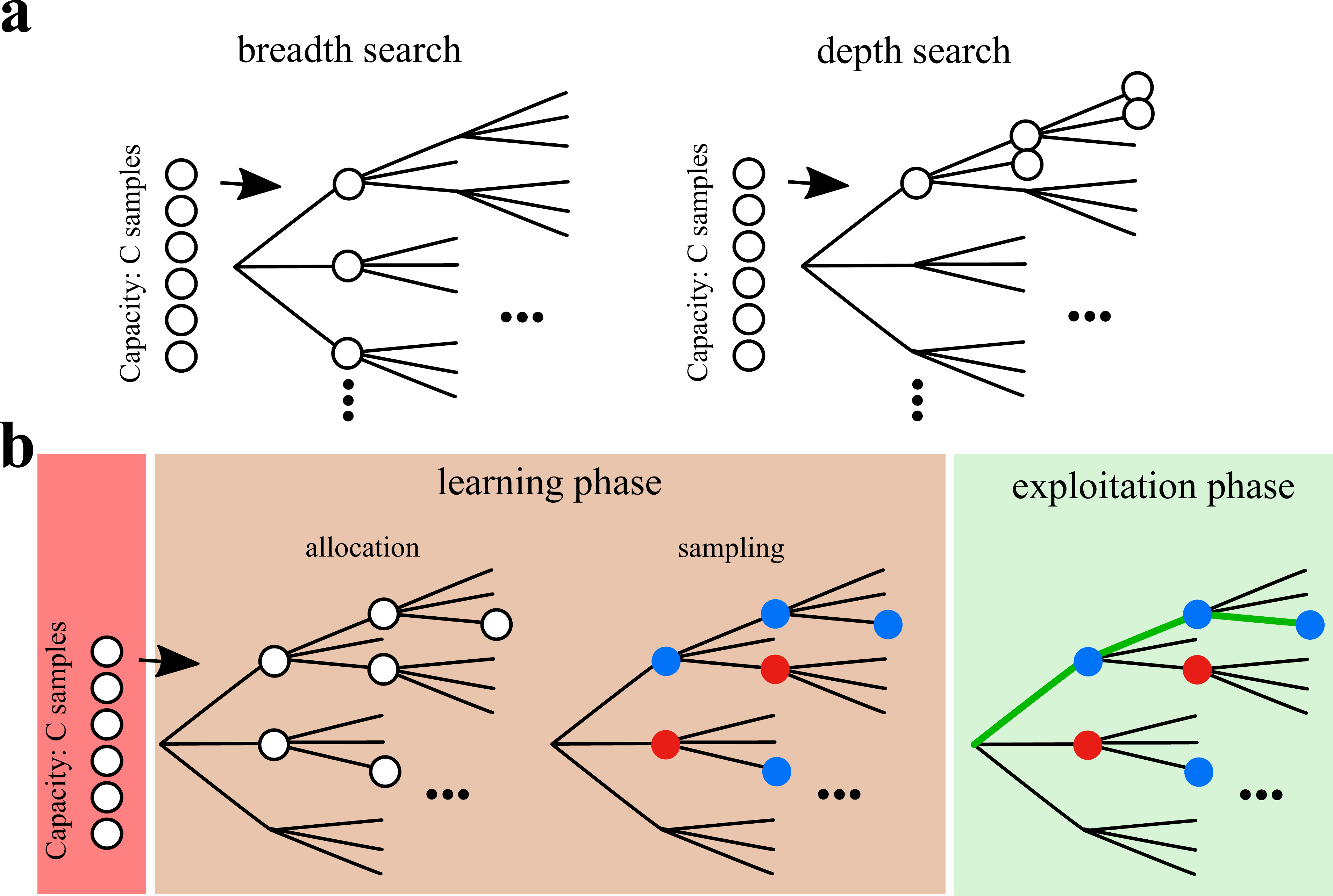} 
\newline
\centering
\caption{Planning decisions in large decision trees with finite sampling capacity. (\textbf{a}) Breadth-depth dilemma in a decision tree of depth $d$ (root is considered level zero) and branching factor $b$ (number of possible actions per node). Nodes correspond to states, and edges correspond to possible actions.  States, if sampled, provide information about whether they deliver on average positive or negative reward. The agent can allocate finite sampling capacity $C$ to gain information about the structure of rewards. Samples can be allocated broadly in the first levels (breadth search, left panel), deeply in few branches (depth search, right panel), or using any intermediate policy  until capacity is exhausted. 
(\textbf{b}) The agent solves the planning problem in two phases: in the exploration or learning phase (brown panel), samples are allocated in one shot to learn about hidden expected rewards of the nodes, and in the exploitation phase (green panel) the learned expected rewards are used to select the optimal path (dark green path).  	
In the example, the 6 samples are allocated (allocation; open circles), after which the agent learns about the rewards from the sampled states (sampling; blue, positive average reward learnt; red, negative average reward learnt). After sampling, the agent can select the optimal sequence of actions. The main problem of the agent is to decide how to allocate samples in the first phase such that the gathered information allows finding a path with the highest possible expected accumulated reward. 
}
\label{fig:fig1}
\end{figure}

\subsection*{A model for search in wide and deep decision trees with finite capacity}

We consider a Markov Decision Process (MDP) that operates in two consecutive phases having different actions (Fig. 1b). The first phase is a learning or exploration phase, while the second one is an exploitation phase. 
In both phases, the underlying structure is a directed rooted tree $\mathcal{G}=(\mathcal{V},\mathcal{E})$ with $d$ levels and homogeneous branching factor, or out-degree, $b$. Thus, each parent node has exactly $b$ children so that there are $b^{k}$ nodes at level $k \in \{0, 1,...,d\}$. Vertices in $\mathcal{V}$ correspond to nodes in the tree, with a total of $|\mathcal{V}|=(b^{d+1}-1)/(b-1)$ of them, and edges $\mathcal{E}$ are links between parents and their $b$ children nodes.  
In the first phase, an action consists of sampling in one shot a subset of $C \leq |\mathcal{V}|-1$ nodes in $\mathcal{G}$ excluding the root node, denoted $\mathcal{V}_{\text{sampled}} \subset \mathcal{V}$, which results in observing the associated random variables $X_s$ for each $s  \in \mathcal{V}_{\text{sampled}}$. Based on the outcomes of the sampled nodes, the agent can update their belief about the expected rewards resulting from visiting them, $R(s)$ for all $s \in \mathcal{V}_{\text{sampled}}$, while the expected reward $R(s)$ resulting from visiting unsampled nodes $s \in \mathcal{V}_{\text{unsampled}}$ remains unchanged. In the second phase, the agent solves a MDP over $\mathcal{G}$, where edges correspond to potential actions, $a \in \mathcal{E} $, and the expected reward resulting from visiting state $s \in \mathcal{V}$ in the tree are the $R(s)$-s updated (or not) in the first phase. Next we describe the above in further detail and provide a rationale for our modeling choices. 

A relevant example is planning a trip to an exotic country: in the first step (root of the trees in Fig. 1) an agent can choose one out of $b$ different countries, from where they can choose one of $b$ different cities to visit in that country, from where they can choose one of $b$ different restaurants, and so on. The planning process can be divided into two phases (Fig. 1b). In the learning phase, the agent learns about what states would be more desirable. In this first phase, actions of the agent do not correspond to actually visiting the nodes $s$ of the tree. Rather, actions correspond to allocating `samples' over certain nodes, resulting in observations that the agent can use to update their belief about the expected reward, $R(s)$, of visiting those nodes. For instance, the agent first gathers information about countries, hotels, etc., by using external (e.g., books) and internal (e.g., memory recollections) information, which results in an update of the expected rewards $R(s)$ resulting from actually visiting those states. This information is used in the exploitation (second) phase to design the best course of actions and commence the trip.

In the learning phase, we assume that the agent has a finite search capacity, modeled as a finite number of samples $C \leq |\mathcal{V}|-1$ that can be allocated over the tree (Fig. 1b, brown panel). The most interesting scenario corresponds to $C \ll |\mathcal{V}|$, when the agent can only sample a small fraction of the nodes in a large decision tree. Thus, the agent's action set equals all possible allocations of the $C$ samples over the graph $\mathcal{G}$ excluding the root node. Formally, every node $s \in \mathcal{V}$ has an associated binary variable $n_s \in \{0,1\}$, indicating whether the node has been sampled, $n_s=1$, or not, $n_s=0$. Note that we assume that nodes can be sampled at most once, and that the finite capacity constraint imposes $\sum_s n_s = C$. Then, the action set can be expressed as  $\mathcal{A}=\{(n_1,n_2,...,n_{|\mathcal{V}|-1}): \sum_s n_s = C , n_s \in \{0,1\} \}$. The nodes with $n_s=1$ define the subset of sampled nodes $\mathcal{V}_{\text{sampled}} \subset \mathcal{V}$. Finite sampling capacity models cognitive and time limitations of the agent, which impedes that a full exhaustive search over all the nodes be possible. 

We assume that the agent allocates all samples at once, that is, without knowing the feedback from the samples. Thus, we consider `one-shot' allocation policies \cite{moreno-bote_heuristics_2020}, which model situations where feedback from the samples arrive with delays longer than the duration of the allocation process (related to explore-then-commit policies \cite{lattimore_bandit_2020}). Many relevant allocation problems are well described by this framework, such as dividing search time to plan a trip, allocating neurons and wiring to different brain areas or cognitive functions during brain development, or dividing budget into several research programs or vaccines. 
One-shot policies are not optimal if the agent is allowed to sample sequentially nodes one by one based on immediate feedback. However, as we show below one-shot optimal policies strongly favor depth over broad search, such that including feedback is expected to further favor depth search, as some tree branches can be pruned early on in the planning process. Therefore, restricting ourselves to one-shot strategies entails a conservative stance to study whether optimal policies favor depth search. 

The result of sampling a node $s$ is to gain information about the expected reward $R(s)$ when visiting the node, which will used in the exploitation phase to optimize the course of actions. We assume that, before sampling starts, the expected reward of any state $s$ is $R(s)=0$. Non-zero average reward can be easily introduced by just adding a constant offset to the rewards independent of the policy. Effectively, we focus on reward excesses compared to a baseline that could result, for instance, from a default policy over which the agent will improve.
Thus, with this definition, if the agent chose a path from the root to the leaves and navigated thought it without having sampled any of the nodes before, the expected accumulated reward associated to such course of actions would be zero. In the trip example, assuming zero expected rewards for all the states before sampling might imply that the agent does not have any initial preference for countries, exotic restaurants and so on. This situation is clearly extreme, as agents might have strong initial, overt preferences \cite{bettman_constructive_1998}. However, strong preferences can only reduce the number of actions to be considered, and therefore will favor depth over breadth policies. Thus, once again our initial no-preference assumption effectively entails a conservative stance. 

When the agent chooses an allocation action $a \in \mathcal{A}$, the graph is partitioned into the sampled and unsampled nodes, $\mathcal{V}_{\text{sampled}}=\{s : n_s = 1\}$ and $\mathcal{V}_{\text{unsampled}}=\{s : n_s = 0\}$ (excluding the root node), respectively. The expected reward of an unsampled node, $n_s=0$, is not updated and thus it remains $R(s)=0$. For a sampled node, $n_s=1$, the belief about its expected reward is updated as follows: we assume that the outcome of sampling the node $s$ is to update $R(s)$ from $0$ to $R_+$ with probability $p$ and to $R_-$ with probability $1-p$, independently for each sampled node (see Fig. 1b, blue and red dots). Thus, $P(R(s)=R_+|n_s=1)=p$ and $P(R(s)=R_-|n_s=1)=1-p$ for a sampled node, and $P(R(s)=0|n_s=0)=1$ for an unsampled node.  
We enforce the condition that the average over updated expected rewards equals zero, that is, $p R_+ + (1-p) R_-=0$, such that sampling a node does not result in net reward or loss. We call this condition the `zero-average constraint', which can be satisfied by taking $R_+=1$ without loss of generality and then using $R_- = -\frac{p}{1-p}$. If the zero-average constraint were not satisfied, we would violate the basic assumption that sampling by itself cannot create or annihilate reward. That is, sampling can change our state of knowledge but not the state or rewards in the world. One way to think about this process is by considering samples as internal `actions' acting over our memory so that they serve to recall or imagine whether some type of food or cities would be desirable \cite{ratcliff_retrieval_1976,shadlen_decision_2016}. Clearly, this process does not change the state or the rewards of the world, although it will be critical to build our preferences. 
It is important to note that the probability of a high reward $p$ in a sampled node measures the overall richness of the environment, and thus how easy is to find a sampled node with positive expected reward $R(s)=R_+$. Therefore, `rich' environments correspond to high $p$ and `poor' environments corresponds to low $p$. 

Once the expected rewards have been updated, the optimal path (Fig. 1b, green path) is computed, which corresponds to the one that has the highest expected accumulated reward based on the observations from the 
samples. Specifically, in the exploitation phase the decision problem forms a standard MDP $\mathcal{M}=(\mathcal{V}, \mathcal{E}, \mathcal{R}, \mathcal{T})$, where states corresponds to nodes in the graph, $s \in \mathcal{V}$, actions correspond to edges of the graph, $a \in \mathcal{E}$, the learned rewards $R(s)$ correspond to the actual expected rewards that result from visiting state $s$, and the transition function $T: (s,a) \rightarrow s'$ between states after an action is made is deterministic. The agent starts in the root node of $\mathcal{G}$, corresponding to the zero-$th$ level, and takes action $a_1 \in \{1,...,b\}$, which results in a deterministic transition to the $a_1-th$ children node $s$ in the first level and the acquisition of a reward with expected value $R(s)$. Recursively, from node $s$ at level $k$, the agent can choose a new action $a_{k} \in \{1,...,b\}$ resulting in a transition to its $a_k-th$ children node $s$ in level $k+1$ and the acquisition of a reward with average $R(s)$. At the $d-th$ level, there are not possible actions and thus leaves correspond to terminal states.
Given the learned expected rewards $R(s)$, the optimal course of actions is found by using backwards induction \cite{sutton_reinforcement_1998}. As we will see, the optimal set of sampled nodes forms a much smaller tree than the original one due to the finite sampling capacity, and then backwards induction over the reduced tree becomes tractable. 

The overall goal of the agent is to determine the best policy to allocate $C$ samples in order to maximize the expected accumulated reward of the optimal path, which implies balancing breadth and depth search: should the agent allocate samples broadly in a few levels, or should allocate few samples per level so that the tree can be sampled deeply?

\subsection*{Value estimation and optimal sample allocations}

We first introduce \emph{exhaustive} allocation policies, which effectively ignore finite capacity by sampling all nodes of a decision tree of depth $d$ and branching factor $b$, but they are simpler to analyze and provide useful tools. 
We then introduce \emph{selective} allocation policies, which allow the agent
to select the number of sampled branches as well as the probability of drawing samples at each tree level under the constraint that the number of allocated samples is on average a fixed capacity $C$. 
As we show below, selective allocations are rich enough to display a broad range of behaviors.  
For each policy we show how to compute its value, defined as the expected accumulated reward of the optimal path. 
To avoid cluttered text, we refer to expected rewards simply as rewards.

\subsubsection*{\emph{Exhaustive allocation}}

An exhaustive allocation policy fully samples all the nodes of a tree with depth $d$ and branching factor $b$. Here, we first compute the probability that an agent can find a path with reward equal to the depth $d$ in such a tree. After this, we calculate the value, $V_{b,d}$, of playing such a tree to develop a useful tool for the case where agents cannot exhaustively sample all nodes. 

We first show that, in general, it is not possible to find a path with all visited nodes having a positive reward. Hence, an optimal path is likely to find a blocked node, that is, a node where all possible actions lead to negative reward, and thus extreme optimism cannot be guaranteed.
By assuming that the reward in a node has value $R_+=1$ with probability $p$ and setting $R_-$ (which is negative) such that the zero-average constraint is satisfied, then the event of finding a path with all positive rewards corresponds to the event that the accumulated reward of the optimal path is the depth $d$ of the tree. We denote the accumulated reward of the optimal path in a tree of depth $d$ by $J_d$, and thus we ask for the probability $P(J_d=d)$. 
If the tree has depth $d=1$ and branching factor $b$, then $P(J_1=1) = 1 - (1-p)^b$. This expression follows from the fact that there are $b$ possible actions, and the probability that none of those actions leads to a reward equal to $R=1$, and thus it is blocked, is $(1-p)^b$.

For $d>1$ we make use of the quantity $Q_{d}=R_{d}+J_{d-1}$, known as action-value, defined as the accumulated reward obtained by first choosing one of the $b$ branches and collect immediate reward $R_d$, and then choosing the best sequence of branches in the remaining $d-1$ levels to collect accumulated reward $J_{d-1}$. 
Note that in principle there are $b$ different action-values $Q_{d}$, one per branch, but as all of them are statistically indistinguishable, an index is not made explicit (the same happens for the rewards $R_{d}$). 
Using this relationship we find

\be
P(J_{d}=d)=1 - \left(1-P(Q_{d}=d)\right)^b = 1-\left(1-pP(J_{d-1}=d-1)\right)^b ,\; d > 1
\label{eq:J_d_max_d}
.
\ee

\noindent
The first equality in Eq. (\ref{eq:J_d_max_d}) comes from the fact that to get an accumulated reward $J_d<d$ it is necessary that none of the $b$ possible actions from the root node leads to $Q_{d}=d$, and that each of those events are statistically independent. The second equality comes from the fact that $P(Q_{d} = d)= pP(J_{d-1}=d-1)$, which is the probability that a particular action from the root node is followed by a state with $R_d=1$, which has probability $p$, and afterwards followed by an optimal path with accumulated reward $d-1$, which has probability $P(J_{d-1}=d-1)$.

We can use the above expression to find cases where the probability of having optimal paths with accumulated reward $d$ approaches zero as $d$ increases. For $b=2$ and $p=\frac{1}{2}$, using Eq. (\ref{eq:J_d_max_d}) we obtain
$P(J_1=1) = \frac{3}{4}$ and $P(J_{d}=d) = 1 - \left(1 - \frac{1}{2} P(J_{d-1}=d-1) \right)^2$ for $d > 1$.
We see that $\lim_{d \rightarrow \infty} P(J_d=d)= 0$, as the only solution to the fixed point equation $P=1-(1-P/2)^2$ is $P=0$. Therefore, the probability that the agent finds a blocking node is one as the tree depth increases. 
For any positive integer $b$ and $p \in [0,1]$, the fixed point equation for large $d$ becomes $1-P = (1-pP)^b$. As the rhs is convex in $P$, positive and has its maximum at $P=0$, the fixed point equation has a non-zero solution only when the rhs' slope at the origin is smaller than $-1$, that is, when $pb>1$. Therefore, if $p$ decreases, then a large enough $b$ ensures a non-zero probability of finding an optimal path with accumulated reward equal to the tree depth. In contrast, if $b \le p^{-1}$, then the probability that the path is blocked with nodes having negative rewards is one.   

After establishing that extreme optimizing is not always guaranteed, we turn to the problem of finding the value of playing the tree with $d$ levels and branching factor $b$, defined as the expected accumulated reward of the optimal paths over such a tree. We provide here the analytical solution for $p=\frac{1}{2}$ and describe the more general analytical solution valid for $p=\frac{1}{n+1}$ and $p=\frac{n}{n+1}$, where $n$ is a positive integer, in Sec.\ref{sec:rational_p} of the Methods. 

For simplicity and without loss of generality we set $R(s)=R_+=1$ and $R(s)=R_-=-1$ with probabilities $p=\frac{1}{2}$, which satisfies the zero-average constraint. Thus, the accumulated reward of a path following a sequence of actions through the tree with $d$ levels can take values $J_d \in \{-d, -d+2, ..., d-2, d\}$. The size of this set is order $\mathcal{O}(d)$, which allows us to compute the value of any tree of depth $d$ in polynomial time. We first compute the probability $P(J_1)$ of the value $J_1 \in \{-1,1\}$ of playing a tree of depth $1$, and then compute the probability $P(J_{d})$ of the value $J_d$ of playing a tree of depth $d$ recursively from $P(J_{d-1})$. Above we showed that $P(J_1=1) = 1 - P(J_1=-1) = 1 - 2^{-b}$ for a tree of depth $1$. Thus, the value of playing such a tree is the average of $J_1$ over sampling outcomes, which equals $V_1 = \mathbb{E}(J_1) = 1 - 2^{1-b}$.

Our algorithm is based on alternating {\em diffusion} and {\em maximization} steps as follows. To find the probability $P(J_{d})$ from $P(J_{d-1})$, we first remind that the action-value $Q_{d}$ is defined as the accumulated reward by taking one action at the root, collect reward $R_d$ and then follow the optimal path in a tree with $d-1$ levels. Written as $Q_{d}=R_d+J_{d-1}$, it has probabilities 

\bea
 && P(Q_{d}=d) = \frac{1}{2} P(J_{d-1}=d-1)
\nonumber
\\
&& P(Q_{d}=d-2) = \frac{1}{2} P(J_{d-1}=d-1) + \frac{1}{2} P(J_{d-1}=d-3)
\nonumber
\\
&& \vdots
\label{eq:dif_step_full}
\\
&& P(Q_{d}=2-d) = \frac{1}{2} P(J_{d-1}=3-d) + \frac{1}{2} P(J_{d-1}=1-d)
\nonumber
\\
&& P(Q_{d}=-d) = \frac{1}{2} P(J_{d-1}=1-d) \;.
\nonumber
\eea

\noindent
This mapping from $P(J_{d-1})$ to $P(Q_{d})$ is a {\em diffusion} step, as the state $J_{d-1}=k$ diffuses to higher, $k+1$, and lower, $k-1$, states of $Q_{d}$ with probability $p=\frac{1}{2}$. We recognize the first identity in Eq. (\ref{eq:dif_step_full}) as the probability that a chosen action followed by the optimal path over a tree with $d-1$ levels leads to an accumulated reward $d$ for the case $p=\frac{1}{2}$, as discussed above.

The diffusion step is followed by the {\em maximization} step, which maps $P(Q_{d})$ into $P(J_{d})$ by

\be
P(J_{d}=k) = P(Q_{d} \le k)^b - P(Q_{d} \le k-1)^b, 
\label{eq:max_step_full}
\ee

\noindent
for $k  \in \{-d, -d+2, ..., d-2, d \}$. Eq. (\ref{eq:max_step_full}) represents a maximization step because the agent will choose the best action out of $b$ available actions, and it expresses that the probability of $J_d=k$ equals the probability of finding at least one action with at most a value of $Q_d=k$. 

In summary, iterating the diffusion and maximization steps in Eqs. (\ref{eq:dif_step_full},\ref{eq:max_step_full}) with initial conditions $P(J_1=1) = 1-P(J_1=-1) = 1 - 2^{-b}$ allows us to compute the value of playing a tree with $d$ levels and $b$ branches by $V_{d,b} = \mathbb{E}(J_{d})$. The number of operations required to determine the value of such a tree is $\mathcal{O}(bd^2)$, as the diffusion step requires $\mathcal{O}(d^2)$ operations due to the presence of $d$ levels and $\mathcal{O}(d)$ different states at each level, and the maximization step involves $\mathcal{O}(b)$ operations for each $J_d=k$ in the calculation of $b$-th powers. In contrast, a direct solution to the problem using dynamic programming requires $\mathcal{O}(b^d)$ operations. This is because the complexity is dominated by the number of nodes in the level before the last one, where there are $b^{d-1}$ nodes, and $b$ operations are needed in each one to solve the $max$ operator before implementing backwards induction. In addition, the complexity of dynamic programming does not take into account the additional need to average over the samples' outcomes, while the diffusion-maximization method in Eqs. (\ref{eq:dif_step_full},\ref{eq:max_step_full}) provides the exact expected value of playing the tree.      

We have studied the value of playing trees as a function of $b$, $d$ and $p$ using the diffusion-maximization method in Eqs. (\ref{eq:dif_step_full},\ref{eq:max_step_full}) for $p=\frac{1}{2}$ and Eqs. (\ref{eq:methods_n_dif_step_full},\ref{eq:methods_n_max_step_full})  and (\ref{eq:methods_1_dif_step_full},\ref{eq:methods_1_max_step_full})  in the Methods for the rational values $p=\frac{n}{n+1}$ and $p=\frac{1}{n+1}$ with positive integer $n$. In all cases, the zero-average constrained is satisfied by setting $R_+=1$ and $R_- = -p/(1-p)$.
The analytical predictions allow us to study very deep trees with, e.g., $d=20$ and $b=5$ at little numerical cost, where the number of nodes is larger than $2 \; 10^{13}$. In contrast, these digits are prohibitive for Bellman - Monte Carlo simulations. 
The value of playing a tree grows monotonically with both its depth and breadth (Fig. 2a), as a tree with a smaller depth or breadth is a subtree that can only have a value equal or smaller than the original tree. Asymptotically, the value grows with unit slope and runs parallel and below the diagonal line (dashed line), which constitutes the highest possible value of any tree, as no tree can have a value above it given our choice $R_+=1$.
With larger $b$, the value runs closer to the diagonal. 
The value of the tree grows monotonically with the probability $p$ of finding high expected reward nodes (Fig. 2b).

\begin{figure}
\includegraphics[width=0.6\textwidth]{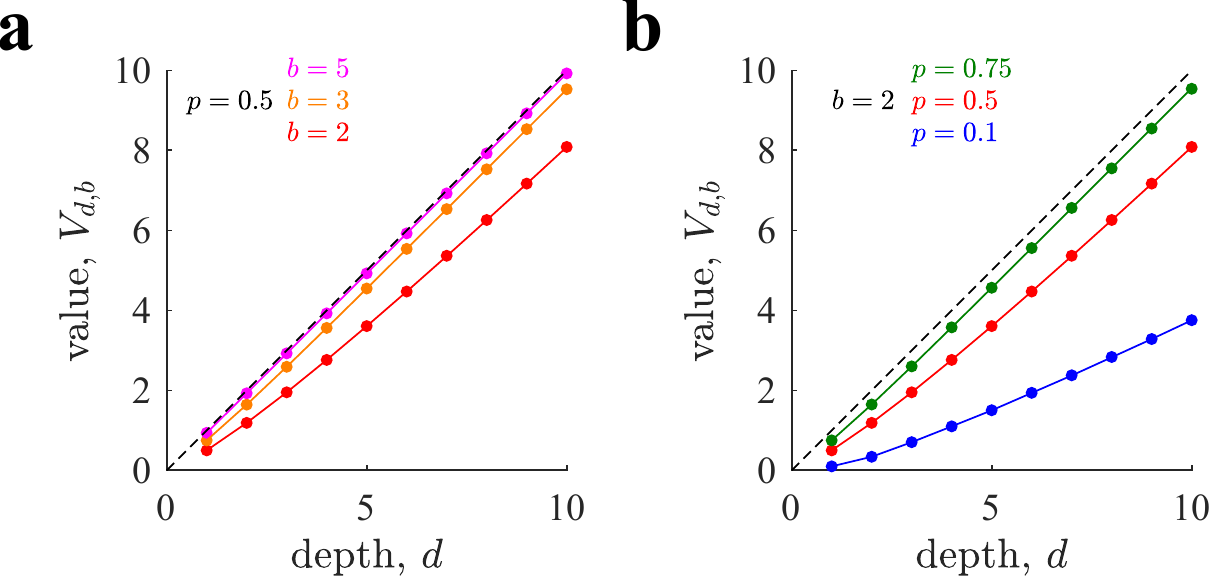} 
\newline
\centering
\caption{
Value $V_{d,b}$ of playing a tree of depth $d$ and branching factor $b$ with exhaustive sampling. 
(\textbf{a}) The value (expected accumulated reward) of playing a tree increases monotonically with both its depth $d$ and its branching factor $b$. In all cases $p=\frac{1}{2}$. For $b=5$ (pink) the value is very close to the maximum possible value (dashed, diagonal line). 
(\textbf{b}) The value of playing the three grows with the probability of high expected reward $p$ in their nodes. In all cases $b=2$.
In both panels, lines correspond to analytical predictions from the diffusion-maximization method, Eqs. (\ref{eq:dif_step_full},\ref{eq:max_step_full}) and Eqs. (\ref{eq:methods_n_dif_step_full},\ref{eq:methods_n_max_step_full},\ref{eq:methods_1_dif_step_full},\ref{eq:methods_1_max_step_full}) (Sec.\ref{sec:rational_p} of the Methods), and dots correspond to Bellman - Monte Carlo simulations (see Sec.\ref{sec:MonteCarlo} of the Methods; average over $10^4$ runs). 
The red lines in the two panels are identical. 
Errors bars are smaller than dots.
}
\label{fig:fig2}
\end{figure}

\subsubsection*{\emph{Selective allocation}}

Now we turn to the central problem of how to optimally sample an infinitely large tree with finite sampling capacity $C$.
Assuming a tree having infinite number of levels and infinite branches per node allows us to consider any possible sampling allocation policy that is solely constrained by finite capacity. 
As such decision tree cannot be sampled exhaustively, we refer to the problem of allocating finite sampling capacity as `selective' allocation. We restrict ourselves to a family of policies where the agent chooses the number of levels $d$ that will be considered as well as the number of branches $b$ per reached node that will be contemplated. Given finite capacity $C$, choosing a large $d$ will imply having to choose a small $b$, thus allowing the agent to trade breadth for depth. 
To provide more flexibility to the allocation policy, we 
also allow that the agent chooses the probability $q_{d-l+1}$ of independently allocating a sample in each node in level $l \in \{1,...,d\}$ (note the reversed order, e.g., $q_1$ refers to the last level $d$). Under this stochastic allocation policy, a node receives a maximum of one sample or can receive none, and thus the allocation is a independent Bernoulli process with sampling probability $q_{d-l+1}$ in each node in level $l$.
Note that here we have relaxed the hard capacity constraint to an average capacity constraint, which turns to be easier to deal with and leads to a smoother analysis. We have observed through numerical simulations that results do not qualitatively differ between hard and average capacity constraints. 

In the following, we first compute the value of sampling a tree of depth $d$ and branching factor $b$ with per-level sampling probabilities $q=(q_1,...,q_d)$. The capacity constraint will be imposed afterwards simply by constraining $d$, $b$ and $q$ to be such that on average the number of allocated samples equals capacity $C$. The algorithm is simply a generalization of the diffusion-maximization algorithm derived for exhaustive allocation in Eqs. (\ref{eq:dif_step_full},\ref{eq:max_step_full}), shown here for the case $p=\frac{1}{2}$ and generalized in Sec.\ref{sec:rational_p} of the Methods to other rational probabilities.  

In contrast to exhaustive allocation, when using selective allocation some nodes might not be sampled, as $q \leq 1$, and thus they will remain having expected reward $R(s)=0$. As before, sampled nodes have values $R(s)=\pm 1$ with probability $\frac{1}{2}$.
Therefore, the value $J_1$ of a depth-$1$ tree is in the set $\{-1,0,1\}$. To compute the expectation of $J_1$ we note that the action-value $Q_1$ of each branch (leaf) has values $\{-1,0,1\}$ with probabilities $P(Q_1=1)=\frac{1}{2} q_1$, $P(Q_1=0)=1-q_1$ and $P(Q_1=-1)=\frac{1}{2} q_1$, which follows from the facts that the node is sampled with probability $q_1$, that if it is sampled then its expected reward $R(s) = \pm 1$ with probability $\frac{1}{2}$, and that if it is not sampled then its expected reward is $R(s) = 0$. As $b$ branches are available each with the same independent distribution of action-values, the value $J_1$ has probabilities $P(J_1=k)=P(Q_1 \le k)^{b}-P(Q_1 \le k-1)^{b}$, which results in $P(J_1=1) = 1- (1-\frac{q_1}{2})^b$, $P(J_1=0) =(1-\frac{q_1}{2})^b - (\frac{q_1}{2})^b $ and $P(J_1=-1) = (\frac{q_1}{2})^b$. 

To compute $P(J_{d})$ recursively from $P(J_{d-1})$, we first relate $P(J_{d-1})$ with $P(Q_{d})$. Since the action-value can be written as $Q_{d}=R_d+J_{d-1}$, where $R_d$ is the reward in a node in level $d$, the diffusion step takes the form 

\bea
&& P(Q_{d}=d) = \frac{1}{2} q_d P(J_{d-1}=d-1)
\nonumber
\\
&& P(Q_{d}=d-1) = (1-q_d) P(J_{d-1}=d-1) + \frac{1}{2} q_d P(J_{d-1}=d-2)
\nonumber
\\
&& P(Q_{d}=d-2) = \frac{1}{2} q_d P(J_{d-1}=d-1) + (1-q_d) P(J_{d-1}=d-2) +  \frac{1}{2} q_d 		P(J_{d-1}=d-3)
\nonumber
\\
&& \vdots
\label{eq:dif_step_q}
\\
&& P(Q_{d}=2-d) = \frac{1}{2} q_d P(J_{d-1}=1-d) + (1-q_d) P(J_{d-1}=2-d) +  \frac{1}{2} q_d 		P(J_{d-1}=3-d)
\nonumber
\\
&& P(Q_{d}=1-d) = \frac{1}{2} q_d P(J_{d-1}=2-d) + (1-q_d) P(J_{d-1}=1-d)
\nonumber
\\
&& P(Q_{d}=-d) = \frac{1}{2} q_d P(J_{d-1}=1-d)
\;.
\nonumber
\eea

\noindent
The diffusion step is followed by the maximization step

\be
P(J_{d}=k) = P(Q_{d} \le k)^{b} - P(Q_{d} \le k-1)^{b}, 
\label{eq:max_step_q}
\ee

\noindent
for $k  \in \{-d, -d+1, ..., d-1, d \}$. 
Iterating the diffusion and maximization steps in Eqs. (\ref{eq:dif_step_q},\ref{eq:max_step_q}) with initial conditions $P(J_1)$ described above
allows us to compute $V_{d,b,q} = \mathbb{E}(J_{d})$, which is the value of playing a tree of depth $d$, branching factor $b$ and per-level sampling probabilities $q$.

We not turn to the problem of optimizing $d$, $b$ and $q$ under the finite capacity constraint. In practice, we can consider a fixed, large $d$ and optimize $b$ and $q$, such that we effectively assume that the sampling probabilities are zero above some depth $d$. If $d$ is large enough this assumption does not impose any restrictions, as the sampling probability can also be zero in levels shallower than the last considered level $d$. 
As the agent is limited by finite sampling capacity, both $b$ and $q$ are constrained by 

\be
C=\sum_{l=1}^d q_{d-l+1} \; b^{l} \;,
\label{eq:C_q_constraint}
\ee

\noindent
which states that the average number of sampled nodes in the subtree must be equal to capacity $C$. 
The optimal $b$ and $q$ are found by

\be
(b^*,q^*) = \argmax_{b,q} V_{d,b,q} \;,
\label{eq:hetero}
\ee

\noindent
subject to the capacity constraint, Eq. (\ref{eq:C_q_constraint}), and for large enough $d$. Optimal allocation policies are numerically found by using a gradient ascent algorithm (Sec. \ref{sec:gradient} of the Methods).
 
In addition to the optimal allocation policies in Eq. (\ref{eq:hetero}), that we call \emph{heterogeneous}, we also consider a subfamily of selective allocations that we call 
\emph{homogeneous}. In a homogeneous allocation policy, the sampling probability is one for all levels except, possibly, the last level, which is chosen to satisfy the finite capacity constraint. As shown below, homogeneous policies are close to optimal and are also simpler to study. 
In a homogeneous selective policy, as in exhaustive allocations, the only choice of the agent is the number of considered branches per reached node $b$.  
Then, effectively, upon choosing $b$, the agent samples $b$ nodes in the first level, and from each of those the agent samples another $b$ nodes in the second level, and so on until capacity is exhausted at some depth $d'\equiv d(b,C)$, that depends on $b$ and $C$. Possibly, not all $b^{d'}$ resulting nodes in the last sampled level $d'$ can be fully sampled. Defining $C_r=C - \sum_{l=1}^{d'-1} q_{d'-l+1} b^{l}$ 
as the remaining number of samples available when reaching the last sampled level $d'$, then each of the $b^{d'}$ considered nodes is given a sample independently with probability $q_1' \equiv q_1(b,C)=C_{r}/b^{d'}$, such that on average total capacity equals $C$.
More specifically, we focus on policies where $b$ is free, $q_1'=C_{r}/b^{d'}$, and $q_2'=...=q_d'=1$ (note again reversed index), with $C_r > 0$. Within this family of allocation policies, the optimal policy is 

\be
b^* = \argmax_b V_{d',b,q'} \;,
\label{eq:homo}
\ee

\noindent
where $V_{d,b,q}=\mathbb{E}(J_{d})$ is found by using the diffusion-maximization method in Eqs. (\ref{eq:dif_step_q},\ref{eq:max_step_q}) and Eqs. (\ref{eq:methods_n_dif_step_q},\ref{eq:methods_n_max_step_q},\ref{eq:methods_1_dif_step_q},\ref{eq:methods_1_max_step_q}) in Sec.\ref{sec:rational_p} of the Methods.

\subsection*{Optimal breadth-depth tradeoffs in allocating finite capacity}

\begin{figure}
\includegraphics[width=0.6\textwidth]{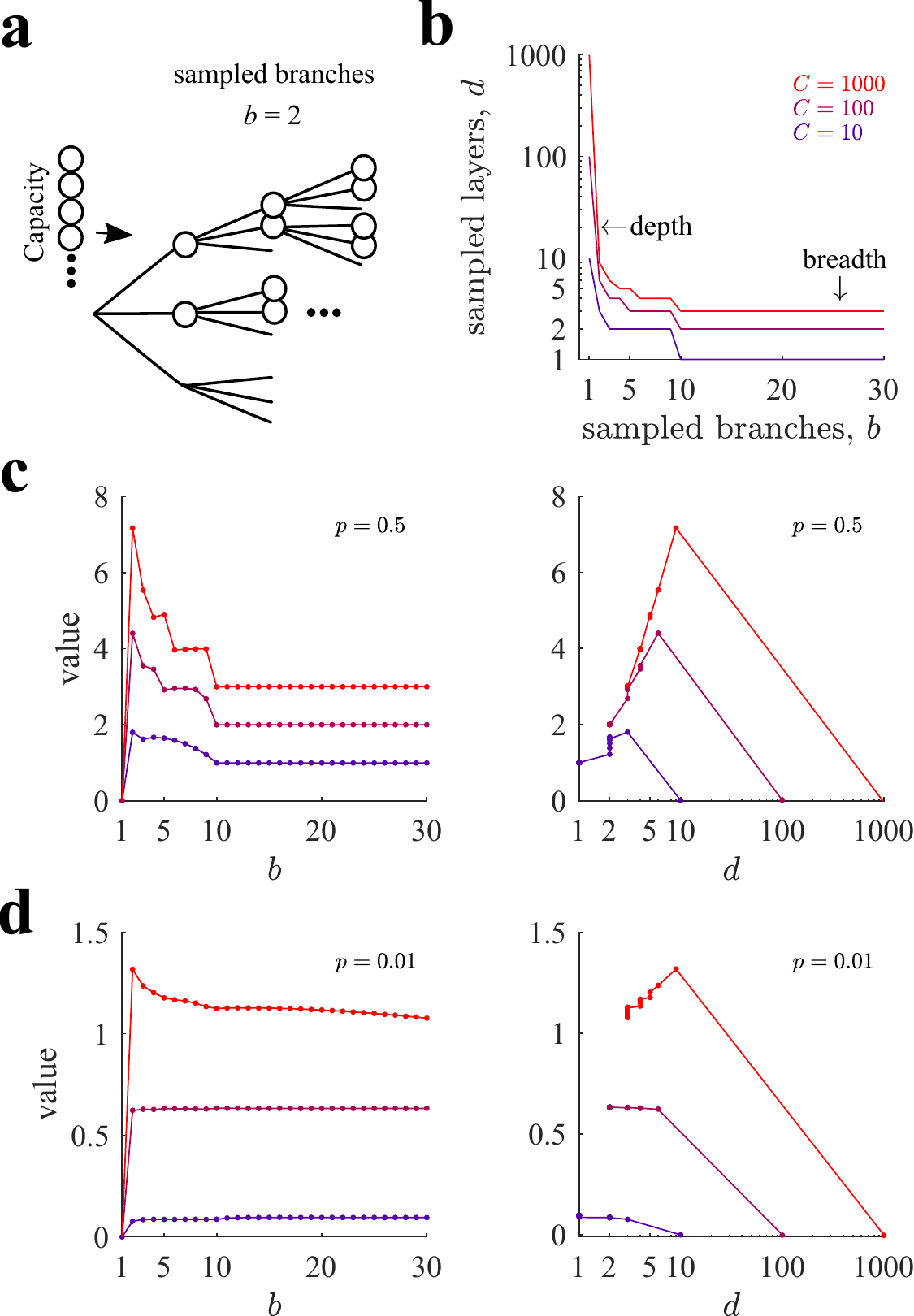} 
\newline
\centering
\caption{
Optimal breadth-depth tradeoffs in sampling decision trees with finite capacity. 	
(\textbf{a}) An agent chooses the number of branches that will be sampled, $b$, per reached node from the root node and continues to sample the tree until capacity is exhausted (\emph{homogeneous} selective allocation). The last nodes are sampled stochastically, so that on average the number of samples equals capacity $C$.
In the example the number of sampled branches is $b=2$.
(\textbf{b}) At fixed capacity, there is a tradeoff between the number of sampled branches and the number of sampled levels. Three values of $C$ have been chosen ($C=10,100,1000$), representing low, medium and high search capacity. For the same number of sampled branches, the number of sampled levels increase with $C$.
The number of sampled levels includes the last level, which might only be partially sampled. Transitions between plateaus occur when the last level is filled up completely with samples. 
(\textbf{c}) Left panel: Value of playing the tree by choosing to sample $b$ branches per reached node with three different values of capacity for $p=\frac{1}{2}$. Note that for each line, selecting $b$ determines the depth of the played tree $d$ (see panel (b)) due to the finite capacity constraint.
The optimal value is attained when the number of sampled branches is $b=2$. 
Right panel: same data as in the right panel are re-plotted as a function of the depth $d$ of the considered subtree. The second longest depth allowed given finite capacity is the optimal allocation to play the tree, which corresponds to $b=2$ in the left panel. The curve shows some vertical jumps because the tree value changes as a function of $b$ even though it does not change $d$.
(\textbf{d}) Same as in panel (c) for $p=0.01$. While at high capacity sampling the tree with a low number of sampled branches remains optimal, at lower capacities it is best to play the tree by favoring breadth over depth. 
In all panels, points correspond to simulations (average over $10^6$ runs) and solid lines correspond to theoretical predictions by Eqs. (\ref{eq:dif_step_q},\ref{eq:max_step_q},\ref{eq:C_q_constraint}) and Eqs. (\ref{eq:methods_1_dif_step_q},\ref{eq:methods_1_max_step_q},\ref{eq:C_q_constraint}) (Sec.\ref{sec:rational_p} of the Methods) for the homogeneous allocation case. 
}
\label{fig:fig3}
\end{figure}

We now describe how optimal selective allocations depend on sampling capacity $C$ and on the richness of the environment as measured by $p$. We start by homogeneous policies, which will be show in the next section to be very close to optimal when compared to heterogeneous policies. 
Selective homogeneous allocations maximize the value of sampling selectively an infinitely broad and deep tree by optimizing the number of sampled branches $b$ (Eqs. \ref{eq:dif_step_q},\ref{eq:max_step_q},\ref{eq:homo}). As capacity is constrained and the sampling probability is one except possibly for the last level, choosing a large $b$ implies reaching shallowly in the tree (Fig. 3b). Thus optimal BD tradeoffs are reflected in the optimal number of considered branches. 
We find that the optimal number of branches is $b^*=2$ for a rich environment ($p=\frac{1}{2}$) regardless of capacity (Fig. 3c, left panel).
Interestingly, we observe that choosing $b=1$ or $b=3$, which are the neighbor policies to the optimal $b^*=2$, leads to a large reduction of performance, indicating that the benefit from correctly choosing the optimum is high. 
The optimal $b^*=2$ favors exploring trees as deep as possible while keeping the possibility of choosing between two branches at each level. 
Indeed, the deepest possible policy resulting from the policy $b=1$ is highly suboptimal (leftmost point in the left panel, and rightmost points in the right panel), as the expected accumulated reward equals zero due to lack of freedom to select the best path.  

For a poor environment (Fig. 3d; $p=0.01$), the optimal number of sampled branches is also $b^*=2$ when capacity is large (peak of red line), but as capacity decreases, $b^*$ increases. Thus, the optimal policy approaches pure breadth at low capacity, which entails exhausting all sampling resources in just the first level. We observe that in this case the dependence of the value of playing the tree with $b$ is very shallow when capacity is small (blue line), and therefore the actual optimal $b^*$ is quite loose.

\begin{figure}
\includegraphics[width=0.7\textwidth]{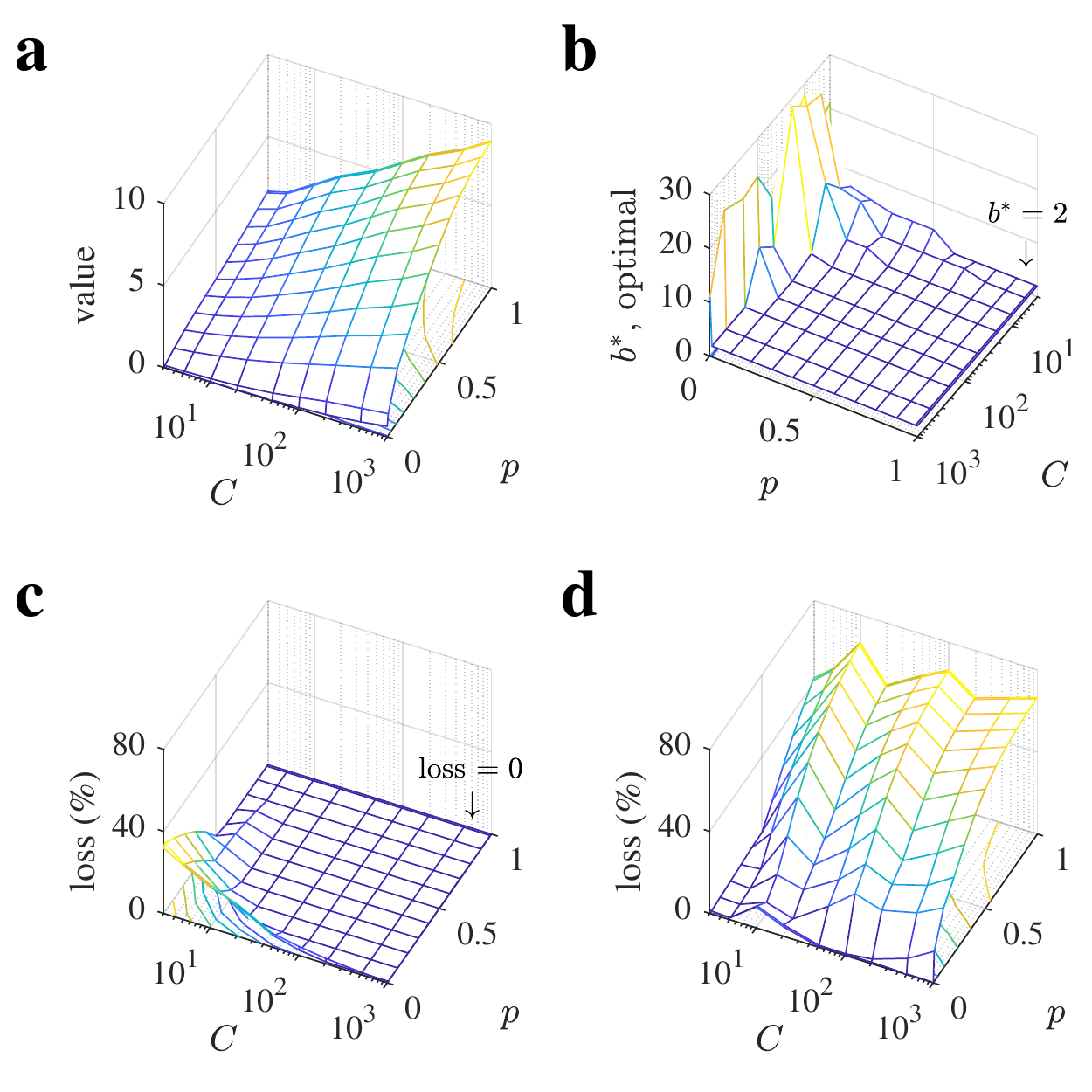} 
\newline
\centering
\caption{
Depth dominates over breadth in large regions of parameter space. 	 	
(\textbf{a}) Value of playing optimally a tree as a function of capacity $C$ and probability $p$.
(\textbf{b}) Optimal number of sampled branches $b^*$ as a function of $C$ and $p$ (note that $C$ and $p$ axes have been rotated for a better data visualization). The large plateau corresponds to the optimal number of sampled branches $b^*=2$. 
(\textbf{c,d}) Loss incurred in playing the tree always with $b=2$ (c), corresponding to depth sampling, or  with $b=20$ (d), corresponding to breadth sampling. The large plateau in panel (c) corresponds to loss equal to zero. Losses are defined as $100 (V_{opt}-V)/V_{opt}$, where $V_{opt}$ is the optimal value (from panel a) and $V$ is the value of sampling the tree with the corresponding heuristic.
Bellman - Monte Carlo simulation results are averaged over $3 \; 10^6$ repetitions. 
}
\label{fig:fig4}
\end{figure}

The results for the two environments described above suggest that depth is always favored when capacity is large enough or whenever the environment is rich, while breadth is only favored at low capacities and for poor environments. 
Further, while optimal breadth policies can be quite loose in that choosing the exact value of $b$ is not very important to maximize value, optimal depth policies are very sensitive to the precise value of the chosen value $b$, always very close to $b=2$, such that variations of it cause large loses in performance. Exploration of a large parameter space confirms the generality of the above results (Fig. 4). In particular, the optimal number of sampled branches is $b^*=2$ for a very large region of the parameters space (Fig. 4b), while an optimal number of branches larger than $2$ mostly occurs exclusively when $p$ is small ($p < 0.1$) or capacity is small ($C<10$). 
If the agent used a depth heuristic consisting in always sampling $2$ branches, then the loss incurred compared to the optimal $b$ would be around $40\%$ at the most, but the region where there are significant deviations in performance concentrates at both low $C$ and $p$ values (Fig. 4c). Indeed, for a very large region of parameter space the loss is zero because almost everywhere the optimal number of sampled branches equals $2$ or because the value of playing the tree is not very sensitivity to $b$. In contrast, using a breadth heuristic where the agent always uses $b=20$ is almost everyone a very poor policy, as losses can reach close to or above $40\%$ in large regions of parameter space (Fig. 4d). Therefore, as an optimal strategy, depth dominates over breadth in larger portions of parameter space, and as a heuristic, depth generalizes much better than breadth. 

Although the optimal policy is quite nuanced as a function of the parameters, a general intuition can be provided about why depth tends to dominate over breadth: exploring a tree allows agents to find paths with accumulated rewards bounded by the length of the path; thus, exploring more deeply leads to knowledge about potentially large rewards excesses as compared to exploring less deeply and following afterwards a default policy. Although this effect seems to be the dominant one, being able to compare among many short courses of actions becomes optimal in poor environments when capacity is small, as it allows securing at least a good enough accumulated reward.

\subsection*{Exploring further into the future is a slightly better policy}

\begin{figure}
\includegraphics[width=0.7\textwidth]{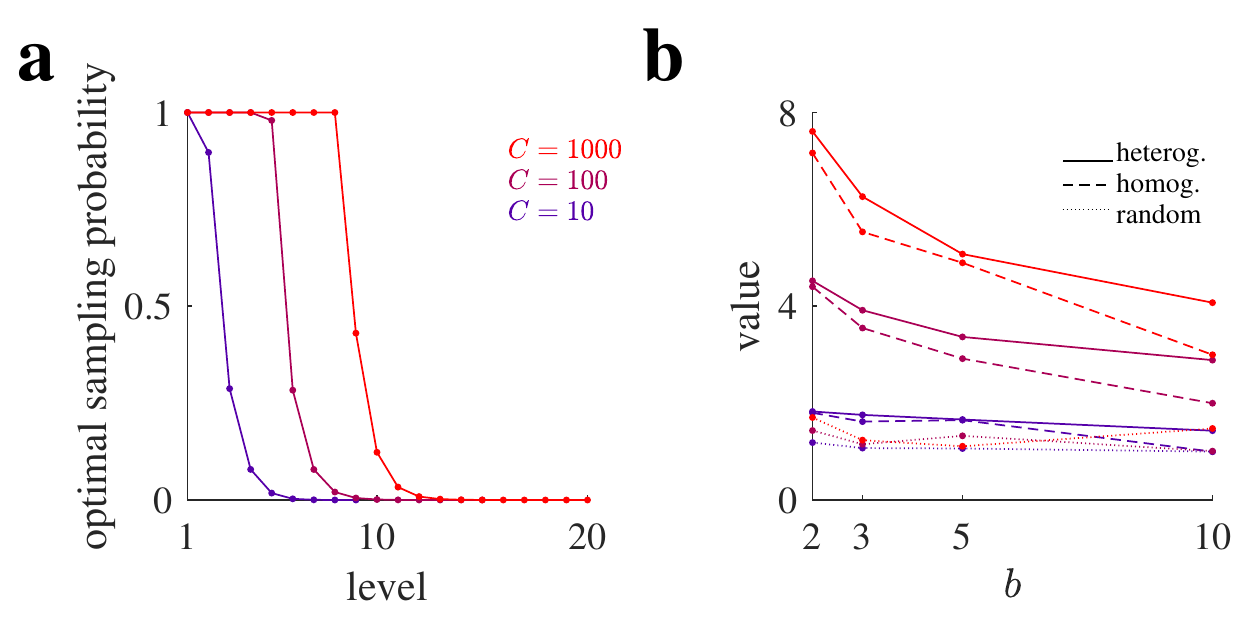} 
\newline
\centering
\caption{
Optimal heterogeneous policies spread samples into the future more deeply than homogeneous policies. 	 	
(\textbf{a}) Optimal sampling probabilities $q$ per level for three capacities and for $b=2$. While for optimal homogeneous policies sampling probabilities equal one except, possibly, for the last level, optimal heterogeneous policies assign non-zero sampling probabilities to deeper levels.  
(\textbf{b}) Value of playing the tree with heterogeneous (full lines), homogeneous (dashed) and random (dotted) policies as a function of the number of considered branches $b$ for three capacities (color code as in previous panel).  
The optimal value is attained when $b=2$ for all cases. Note that optimal values for homogeneous policies are below but very close to the optimal values of heterogeneous policies.
For heterogeneous and random policies, we limit the number of considered levels somehow arbitrarily to $d = 2  \lfloor ln(C)/ln(b) \rfloor + 3$, where $\lfloor x \rfloor$ is the floor function, which allows in a simple way agents to spread samples, if optimal, well beyond the sampled levels by homogeneous policies. Random policies allocate samples with the same probability to every node of the tree of depth $d$ and also satisfy the finite capacity constraint, Eq. (\ref{eq:C_q_constraint}).
Optimal policies and values for heterogeneous and for homogeneous selective allocations are computed using Eqs. (\ref{eq:dif_step_q},\ref{eq:max_step_q},\ref{eq:hetero}) and Eqs.  (\ref{eq:dif_step_q},\ref{eq:max_step_q},\ref{eq:homo}) for $p=\frac{1}{2}$, respectively, inside a gradient ascent (see Sec.\ref{sec:gradient} of the Methods). For different $p$ results are similar.   
}
\label{fig:fig5}
\end{figure}

One important question is how much can be gained by giving to the agent a larger degree of flexibility in allocating samples over the levels. In heterogeneous selective policies, the agent is free to choose the number of branches to be considered as well as the sampling probabilities for each of the levels (Eqs. \ref{eq:dif_step_q},\ref{eq:max_step_q},\ref{eq:hetero}). Therefore, in contrast to homogeneous selective policies, the agent can decide not to allocate samples to the first levels and reserve them for deeper levels. Our analysis, however, shows that it is not the best allocation policy, as optimal heterogeneous policies sample exhaustively the first levels, as homogeneous policies do (Fig. 5a). 
One important difference is that optimal heterogeneous policies explore further into the future than homogeneous policies. 
This is accomplished by using sampling probabilities decaying to zero in the last few sampled levels. This is in contrast to homogeneous policies, where only the last level is given, possibly, a sampling probability smaller than one. 
Thus, exploring slightly further into the future provides a surplus value of playing the tree (Fig. 5b, full lines), but it is only marginally better than the one obtained from homogeneous policies (dashed lines), which are much simpler to implement due to their fixed sampling probability structure. As in the case of homogeneous policies, heterogeneous policies attain their optimal value when the number of considered branches is $2$, thus favoring depth over breadth search. 
Finally, we tested random policies where samples are allocated with the same probability to the nodes of the first layers of the tree until capacity is exhausted (dotted lines), and found that they are much worst than the optimal policies.

\section*{Discussion}

Agents with limited resources face breadth-depth tradeoffs when looking for the best course of actions in deep and wide decision trees. To gain information about the best course, an agent might allocate resources to sample many actions per level at the cost of not exploring the tree deeply, or allocate resources to sample deeply the tree at the risk is missing relevant actions. We have found that deep imagination is favored over breadth in a broad range of conditions, with very little balance between the two: it is almost always optimal to sample just a couple of actions per depth level such that the tree is explored as deeply as possible while sacrificing wide exploration. In addition, using depth as a heuristic for all cases incurs much smaller errors than assuming a breadth heuristic. We have provided analytical expressions for this problem, which allows us to study the optimal allocations in very large decision trees. 

During planning, we very often picture the course of actions as an imaginary episode, from taking the plane to visiting the first museum, in a process that has been called imagination-based planning, model-based planning, mental simulations or emulation, each term carrying somehow different meanings \cite{clark_towards_1999,grush_emulation_2004,nanay_role_2016,doll_ubiquity_2012,simons_brain_2017,pezzulo_coordinating_2008,hamrick_analogues_2019}. 
Imagination strongly affects choices through the availability of the imagined content \cite{tversky_availability:_1973}, and it is used when the value of the options are unknown and thus preferences need to be built on the fly \cite{nanay_role_2016}. However, imagination-based planning is slow and there is no evidence that can run in parallel \cite{gupta_hippocampal_2010,pfeiffer_hippocampal_2013}, implying that as an algorithm for exploring deep and wide decision trees it might not be efficient. Indeed, very few courses of actions ($\sim 5-10$) are considered in our `minds' before a decision is made \cite{hauser_evaluation_1990-1,stigler_economics_1961,roberts_development_1991-1,mehta_price_2003-1,santos_testing_2012-1,scheibehenne_can_2010-1}, and in some cases the imagined episodes can be characteristically long, like when playing chess \cite{simon_theories_1972}, although their depth can be adapted to the current constraints and time pressure \cite{keramati_adaptive_2016}.
As an alternative to its apparent clumsiness, \emph{deep imagination} --the process of sampling few long sequences of states and actions-- might have evolved as the favored solution to breadth-depth tradeoffs in model-based planning under limited resources against policies that sample many short sequences. Our results provide a theoretical foundation for the optimality of deep imagination in model-based planning by showing that it becomes the dominant strategy in one-shot allocations of resources over a broad range of capacity and environmental parameters. 
Recent deep-learning work has studied through numerical simulations how agents can benefit from imagining future steps by using models of the environment \cite{hamrick_metacontrol_2017,pascanu_learning_2017,weber_imagination-augmented_2018,hafner_dream_2020}, and thus our results might help to clarify and stress the importance of deep tree sampling through mental simulations of state transitions. 

Deep imagination resembles depth-first tree search algorithms in that they both favor deep over broad exploration \cite{pearl_search_1987,korf_depth-first_1985}. However, depth-first search starts by sampling deeply until a terminal state is found, but actually reaching a terminal state in very deep trees can be unpractical \cite{browne_survey_2012} and even the notion of terminal state might not be well-defined, as in continuing tasks \cite{sutton_reinforcement_1998}. In very deep decision trees such strategy would imply the sampling of a single course of actions until exhaustion of resources, which is a highly suboptimal strategy, as we have shown (see Fig. 3 with $b=1$). 
Another family of search algorithms, called breadth-first search \cite{korf_depth-first_1985}, and other approaches that give finite sampling probability to every action at each visited nodes, such as Monte Carlo tree search \cite{browne_survey_2012} or $\epsilon$-greedy reinforcement learning methods \cite{sutton_reinforcement_1998}, poorly scale when the branching factor of the tree is very large, and thus they are unpractical approaches for BD dilemmas.
In contrast, deep imagination samples two actions per visited node until resources are exhausted, which allows selecting the best among a large number of paths, and at the same time constitutes an algorithm that is simple to implement and generalizes well.
Due to finite capacity, any algorithm can only sample a large decision tree up to some finite depth, 
which leaves open the question of how the agent should act afterwards. Following the approach of plan-until-habit strategies \cite{keramati_adaptive_2016,sezener_optimizing_2019}, we have assumed that agents can follow a random, or default, strategy after the last sampled level of the tree, 
such that different allocations policies with different sampled depth and branching factors could be compared on an equal footing. 

One important assumption in our work is the one-shot nature of the sample allocation. 
Many important decisions have delayed feedback, like allocating funding budget to vaccine companies, choosing college, or planning a round of interviews for a faculty position, and thus they are well modeled as one-shot finite-resource allocations \cite{hauser_evaluation_1990-1,roberts_development_1991-1,mehta_price_2003-1}. 
However, other decisions involve quicker feedback and then the allocation of resources could be adapted on the fly. 
Although our results are yet to be extended to sequential problems where at every step a compound action is to be made, 
we conjecture that such extension will not substantially change the close-to-optimality of deep sampling, although a bias towards more breadth is expected \cite{moreno-bote_heuristics_2020}.
Further, pre-computing allocation strategies at design-time and using them afterwards might lift up the burden of performing heavy online computations that would require complex tree expansion in large state spaces. 
Thus, by hard-wiring these strategies much of the overload caused by meta-reasoning \cite{russell_principles_1991,gershman_computational_2015,griffiths_rational_2015}
could be alleviated, allowing agents to use their finite resources for the tasks that change on a faster time scale. 
Finally, it is important to note that, in contrast to many experimental frameworks on binary choices or very low number of options \cite{gold_neural_2007,churchland_decision-making_2008,drugowitsch_cost_2012,krajbich_visual_2010-1} and games \cite{simon_theories_1972,krusche_adaptive_2018} where the number of actions is highly constrained by design, realistic decisions face too many immediate options to be all considered
\cite{hauser_evaluation_1990-1,stigler_economics_1961,roberts_development_1991-1,scheibehenne_can_2010-1}, and thus a first decision that cannot be deferred is how many of those to focus on in the first place \cite{moreno-bote_heuristics_2020,ramirez-ruiz_optimal_2021,krajbich_visual_2010-1,hayden_neuronal_2018}. All in all, the optimal BD tradeoffs that we have characterized here might play an important role even in cases that substantially depart from our modeling assumptions.

In summary, we have provided a theoretical foundation for deep imagination as a close to optimal policy for allocating finite resources in wide and large decision trees. Many of the features of the optimal allocations that we have described here can be tested by controlling parametrically the available capacity of agents and the properties of the environment \cite{ramirez-ruiz_optimal_2021} by using similar experimental paradigms to those recently developed \cite{vidal21}, which constitutes a relevant future direction.

\section*{Methods}

\subsection{Bellman - Monte Carlo simulations}
\label{sec:MonteCarlo}

The exact values of playing tree for a subset of rational values of $p$ are computed using the diffusion-maximization algorithm. 
For probabilities of positive rewards $p$ not in that set, we can estimate the value by Bellman - Monte Carlo simulations. We first sample each node in the tree (except the root node) to determine the reward associated with it, $R(s)$, which is $R(s)=R_+$ with probability $p$ and $R(s)=R_-$ with probability $1-p$. We take $R_+=1$ and $R_- = -p/(1-p)$ to satisfy the zero-average constraint. Based on the learned $R(s)$-s, we compute the value of the tree by using backwards induction from the last nodes until reaching the root node. Specifically, the leaf nodes have value $V(s)=R(s)$. Recursively, going backwards, the value of a node $s$ at depth $m$ is computed from the values of its children nodes $s' \in \text{ch}(s)$ at depth $m+1$ as $V(s) = \max_{s' \in \text{ch}(s)} (R(s') + V(s') )$. The value of playing the tree with the specific realization of the $R(s)$-s is the value of the root node computed that way. The value of playing the tree is the average value over a large number of realizations of the $R(s)$-s, as indicated in the corresponding figures.

\subsection{Gradient ascent}
\label{sec:gradient}

For each $b$ we optimize $q$ in Eq. (\ref{eq:hetero}) under the capacity constraint, Eq. (\ref{eq:C_q_constraint}), by a gradient ascent method. 
The unconstrained gradient of the value $V_{b,d,q}$ is numerically computed for an initial $q$
using a discretization step size $\Delta q_k=10^{-7}$, $k \in \{1,...,d \}$.
The unconstrained gradient is then projected onto the capacity constraint plane defined by Eq.  (\ref{eq:C_q_constraint}). Then, the projected gradient multiplied by a learning rate $\eta=10^{-3}$ is added to the original $q$, from where a new $q$ is proposed. If the resulting $q$ has a component $q_k$ that does not satisfy the constraint $0 \leq q_k \leq 1$, then $q_k$ is moved to either $0$ or $1$, whichever is closer. This movement can make $q$ in turn to be outside the capacity constraint plane, so a new projection onto the constraint plain is performed. The projections and movements are repeated until $q$ satisfies both constraints, leading to a new valid $q$.
From the new $q$, an unconstrained gradient is computed again, and the procedure continues up to a maximum of $10^6$ iterations or when the improvement in the value $V_{b,d,q}$ is less than a tolerance of $10^{-9}$. To avoid numerical instabilities for very deep trees ($d>50$), the probabilities $P(J_d)$ are normalized to sum one at every iteration. One order of magnitude differences in the ranges of step sizes, learning rates and tolerances, and all tested initial conditions for $q$ give almost identical results to those reported in the main text.

\subsection{Value of exhaustive or selective search in a large tree with rational  \texorpdfstring{$p$}{p}}
\label{sec:rational_p}

We extend our results for $p=\frac{1}{2}$ to the case of rational values $p=p_+=\frac{n}{n+1}$ and $p=p_+=\frac{1}{n+1}$ for any positive integer $n$. The zero-average reward constraint enforces that $p_++p_-=1$ and $p_+R_++p_-R_-=0$.
We arbitrarily take $R_+=1$ and select $R_-$ so that the zero-average reward constraint is satisfied.

\subsubsection{Reward probability \texorpdfstring{\boldmath$p=\frac{n}{n+1}$}{n}}

We first consider $p=p_+=\frac{n}{n+1}$, which implies $p_-=\frac{1}{n+1}$.
The zero-average constraint results in $R_-=-n$.
We describe below how to compute the value of playing a large tree exhaustively and selectively with such a probability $p$ of positive reward. 

\paragraph{\emph{Exhaustive allocation}.}

We begin by describing the value of a tree with one level ($d=1$), which will serve as initial condition for the diffusion-maximization algorithm. In this case, the accumulated reward can only be $1$ or $-n$, that is, $J_1 \in \left\{1, -n \right\}$. Thus

\begin{equation*}
    P\left(J_1=1\right)=1-P\left(J_1=-n\right)=1-\left(\frac{1}{n+1}\right)^b\;,
\end{equation*}
where $b$ is the number of branches. 

As we have seen for $p=\frac{1}{2}$ in the main text, we can compute the probabilities for a tree of depth $d$ starting from the probabilities of the accumulated reward of a tree of depth $d-1$ by alternating the \emph{diffusion} and \emph{maximization} steps.
The diffusion step uses the probabilities of the accumulated reward $J_{d-1}$ of a tree of depth $d-1$ to compute the action values $Q_{d}$ of a tree of depth $d$ using the possible rewards $R_d=\left\{R_+=1,R_-=-n\right\}$. 
Both the accumulated rewards $J_d$ and the action values $Q_d$ for a tree of depth $d$ can take values $k=-nd + (n+1)i$, with $i\in\left\{0,1,2,\dots,d\right\}$, where $i$ is number of times the positive reward $1$ was observed in the best possible path. 

Using the above, the diffusion step becomes

\begin{equation}
    \begin{split}
        &P\left(Q_d=k\right)=\frac{1}{n+1}P\left(J_{d-1}=k+n\right)+\frac{n}{n+1}P\left(J_{d-1}=k-1\right)\;,\\
    \end{split}
    \label{eq:methods_n_dif_step_full}
\end{equation}

\noindent
where it is understood that $P(J_{d-1}=k')=0$ if $k'$ lies outside the domain of $J_{d-1}$, in particular when $k'> d-1$ or $k'< -n(d-1)$, and thus some terms in the rhs of the above equation can become zero, by definition.

The maximization step is, as before,
\begin{equation}
    P\left(J_d=k\right)=\left(P\left(Q_d\le k\right)\right)^b-\left(P\left(Q_d\le k-1\right)\right)^b\;.
     \label{eq:methods_n_max_step_full}
\end{equation}

\paragraph{\emph{Selective allocation}.}

The average finite capacity constraint enforces that
\begin{equation*}
    C=\sum_{l=1}^d q_{d-l+1}b^l\;,
\end{equation*}
where $q_{d-l+1}$ is the sampling probability of tree level $l$.
We underline the reverse order of the index of $q$, which is due to the fact that we are describing a backward algorithm: $q_1$ will appear in the first step and corresponds to the last level, $q_2$ in the second step and corresponds to the second last level, and so on. 
In selective allocation of samples, it is possible that a node is not sampled, and thus the possible values of both $J_d$ and $Q_d$ are

\begin{equation*}
    k=i - n j\;,
\end{equation*}
with $i,j\in\{0, 1 \dots, d\}$ and $i+j\le d$, where $i$ is the number of times the positive reward $1$ is observed, and $j$ is the number of times the negative reward $-n$ is observed. 

We now proceed to compute the value of a tree with one level, and then use the diffusion-maximization algorithm to compute the value of a tree with any arbitrary depth $d$. The probabilities of the action values $Q_1$ for the branches of such a tree are
\begin{align*}
    P\left(Q_1=-n\right)&=q_1p_-=\frac{q_1}{n+1}\\
    P\left(Q_1=0\right)&=1-q_1\\
    P\left(Q_1=1\right)&=q_1p_+=\frac{n q_1}{n+1}\;,
\end{align*}
and by using the maximization step, we obtain that the values $J_1$ take probabilities 
\begin{align*}
    P\left(J_1=-n\right)&=\left(P\left(Q_1\le -n\right)\right)^b\\
    P\left(J_1=0\right)&=\left(P\left(Q_1\le 0\right)\right)^b- \left(P\left(Q_1\le -n\right)\right)^b\\
    P\left(J_1=1\right)&=\left(P\left(Q_1\le 1\right)\right)^b- \left(P\left(Q_1\le 0\right)\right)^b\;.
\end{align*}

Now, the diffusion step is
\begin{equation}
P\left(Q_d=k\right)
=\left(1-q_d\right)P\left(J_{d-1}=k\right)+
    \frac{1}{n+1}q_dP\left(J_{d-1}=k + n\right)+
    \frac{n}{n+1}q_dP\left(J_{d-1}=k -1\right)\;,
\label{eq:methods_n_dif_step_q}
\end{equation}
where, again, it is understood that $P(J_{d-1}=k')=0$ when $k'$ lies outside the domain of $J_{d-1}$, in particular when $k'> d-1$ or $k'< -n (d-1)$, and thus many terms contribute zero. 

The diffusion step is then followed by the usual maximization step
\begin{equation}
    P\left(J_d=k\right)=\left(P\left(Q_d\le k\right)\right)^b-\left(P\left(Q_d\le k-1\right)\right)^b\;.
    \label{eq:methods_n_max_step_q}
\end{equation}

\subsubsection{Algorithmic complexity}
The complexity of the algorithm is proportional to the number of equations, which equals the sum of the number of possible different states per level. As we said above, the possible state values $J_s$ at level $s$ are $k=i-nj$, with $i,j \ge 0$ and $i+j \le s$. As $n$ is an integer, it is possible to have repeated values of $k$ for different values of $i$ and $j$ within the allowed set. 

\begin{figure}[h!]
 \begin{subfigure}{0.45\textwidth}
         \centering
         \includegraphics[scale=0.26]{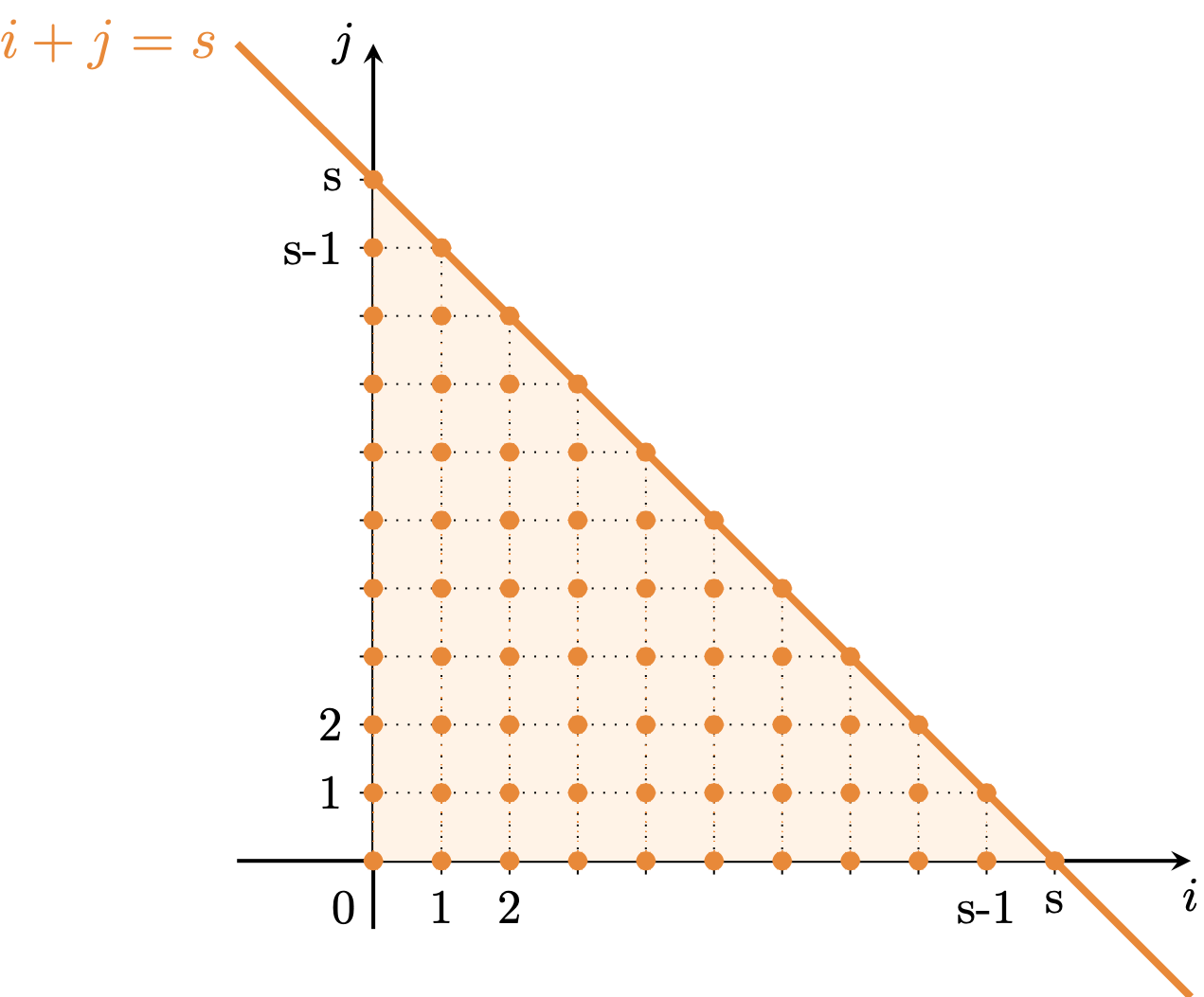}
         \caption{}
             \label{fig:ij}
     \end{subfigure}%
     \hfill
     \begin{subfigure}{0.45\textwidth}
         \centering
         \includegraphics[scale=0.26]{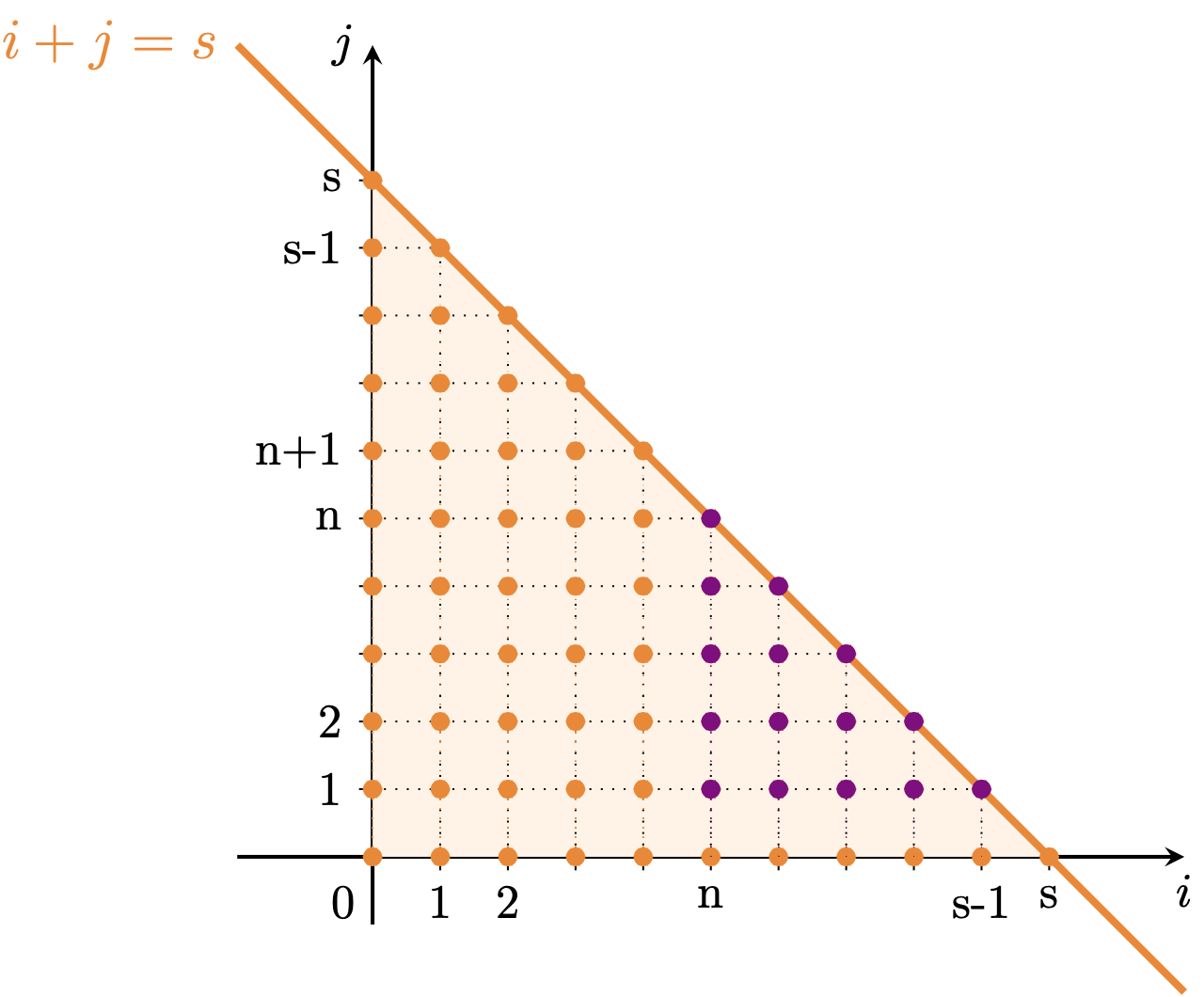}
             \caption{}
                 \label{fig:ij_overlap}
                 
     \end{subfigure}
     \caption{(a) Possible $(i,j)$ pairs for admissible states at layer $s$. (b) Same as in (a); in purple, the pairs leading to an already considered state (overlapping states).}
\end{figure}

To count the number of distinct states, we start by noticing that if $j=0$, then $k=i$, and thus there are $s+1$ distinct states (Fig. (\ref{fig:ij}), orange points in the bottom row of the triangle). 
Assume first that $s<n$. 
If $j=1$, then $k=i-n$, where $i$ lies between $0$ and $s-1$ (second bottom row of points in the triangle).
As $s<n$, the resulting states $k=i-n$ do not reach $k=0$, and thus all of them are distinct from those corresponding to the bottom row. 
If $j=2$, the states are $k=i-2n$, where $i$ lies between $0$ and $s-2$ (third bottom row), and as the values of $k$ do not reach $-n$, the new states are all new. 
In conclusion if $s<n$ the total number of distinct states $N(n,s)$ in level $s$ is

\begin{equation*}
    N(n,s) = \frac{(s+1)(s+2)}{2}\;,
\end{equation*}

For $s \ge n$, there are many values of $i$ and $j$ that result in repeated states $k$ (Fig. (\ref{fig:ij_overlap}), violet points).
If $j=0$, then $k=i$, resulting in $s+1$ distinct states, as before (orange points in the bottom row of the triangle). 
If $j=1$, then $k=i-n$, resulting in the states $\{-n,n+1,...,0,...,s-n\}$, of which all states equal or above $0$ are repeated (violet points in the second bottom row). Thus, there are $n$ new states.
Extending the above, for each $j$ in $\{1,...,n\}$ there are $n$ new states, and for larger values of $j$ the new states are $s-j+1$.

In conclusion, if $s \ge n$ the total number of distinct states $N(n,s)$ in level $s$ is

\begin{equation*}
    N(n,s) = (n+1)s - \frac{n(n-1)}{2} + 1\;,
\end{equation*}

From here, the scaling of states is proportional to the level $s$, and for large $s$ the term $ns$ dominates. Therefore, when summing up distinct states from the first to the last level $d$ of the tree, we conclude that the complexity of the maximization-diffusion algorithm is
$\mathcal{O}\left(n d^2 b\right)$, where we take into account that for every state we need to perform a maximization step (a power operation that counts $b$ per state).  
Analogous steps can be made for the case considered next of $p=\frac{1}{n+1}$ to reach to an identical algorithmic complexity.

\subsubsection{Reward probability \texorpdfstring{\boldmath$p=\frac{1}{n+1}$}{uno}}
 
We proceed by considering $p=p_+=\frac{1}{n+1}$ which implies $p_-=\frac{n}{n+1}$. The zero-average reward leads in this case to a negative reward $R_-=-\frac{1}{n}$. We show here how to compute the value of playing a large tree, exhaustively and selectively, and with such reward probability $p_+$.

\paragraph{\emph{Exhaustive allocation}.}
As shown before, the initial conditions for the diffusion-maximization algorithm come from the value of a tree with just one level $(d=1)$. For a single level tree the accumulated reward can only be $1$ or $-\frac{1}{n}$, namely $J_1\in\{1,-\frac{1}{n}\}$. Thus, for a number $b$ of branches 
\begin{equation*}
    P\left(J_1=1\right)=1- P\left(J_1=-\frac{1}{n}\right) =1-\left(\frac{n}{n+1}\right)^b\;.
\end{equation*}

Again we can compute the probabilities of $J_d$ for a tree of depth $d$ from the probabilities of $J_{d-1}$ for a tree of depth $d-1$ using \emph{diffusion}-\emph{maximization}. In the diffusion step, we use the probabilities of $J_{d-1}$ of a tree of depth $d-1$ to compute the action values $Q_d$ of the tree of depth $d$ along with the possible rewards $R_d=\{R_+=1, R_-=-\frac{1}{n}\}$.
For a tree of depth $d$, both the accumulated reward $J_d$ and the action value $Q_d$ can take the values $k=-\frac{d}{n}+\left(\frac{1}{n}+1\right)i$ with $i\in\{0,1,\dots,d\}$, where $i$ is the number of times that the positive reward $R_+=1$ is observed. 

Now, the diffusion step becomes 
\begin{equation}
 P\left(Q_d=k\right)=\frac{n}{n+1}P\left(J_{d-1}=k+\frac{1}{n}\right)+\frac{1}{n+1}P\left(J_{d-1}=k-1\right)\;,
  \label{eq:methods_1_dif_step_full}
\end{equation}
where again the probabilities $P(J_{d-1}=k')$ are zero when $k'$ lies outside the domain of $J_{d-1}$, in particular when $k'> d-1$ or $k'<-\frac{d-1}{n}$.

After the diffusion, the maximization step is always
\begin{equation}
    P\left(J_d=k\right)=\left(P\left(Q_d\le k\right)\right)^b-\left(P\left(Q_d\le k-1\right)\right)^b\;.
     \label{eq:methods_1_max_step_full}
\end{equation}

\paragraph{\emph{Selective allocation}.}
As we have shown in the main text for $p=\frac{1}{2}$, and previously here for $p=\frac{n}{n+1}$, in selective allocation we consider the average finite capacity constraint

\begin{equation*}
    C=\sum_{l=1}^d q_{d-l+1}b^l\;,
\end{equation*}
where $q_{d-l+1}$ is the sampling probability of tree level $l$.
As nodes might not be sampled, the possible values of both $J_d$ and $Q_d$ are
\begin{equation*}
    k = i-\frac{j}{n}\;,
\end{equation*}
with $i,j\in\{0,1,\dots, d\}$ and $i+j\le d$, where $i$ is the number of times that the positive reward $1$ is observed, and $j$ is the number of times that the the negative reward $-\frac{1}{n}$ is observed in the best possible path. 
We first compute the value of a tree with depth $1$ and then use the diffusion-maximization algorithm to perform induction over $d$.
The probabilities of the action values $Q_1$ for the branches of a tree with $d=1$ are

\begin{align*}
    P\left(Q_1=-\frac{1}{n}\right)&=q_1p_-=\frac{nq_1}{n+1}\\
    P\left(Q_1=0\right)&=\left(1-q_1\right)\\
    P\left(Q_1=1\right)&=q_1p_+=\frac{q_1}{n+1}\;.\\
\end{align*}
Thus, the probability of $J_1$ are obtained by using the maximization step
\begin{align*}
    P\left(J_1=-\frac{1}{n}\right)&=\left(P\left(Q_1\le -\frac{1}{n}\right)\right)^b\\
    P\left(J_1=0\right)&=\left(P\left(Q_1\le 0\right)\right)^b- \left(P\left(Q_1\le -\frac{1}{n}\right)\right)^b\\
    P\left(J_1=1\right)&=\left(P\left(Q_1\le 1\right)\right)^b- \left(P\left(Q_1\le 0\right)\right)^b\;.
\end{align*}

Given these initial conditions, it is easy to see that the diffusion step for level $d$ is 

\begin{equation}
   %
   P\left(Q_d=k\right)
   =\left(1-q_d\right)P\left(J_{d-1}=k\right)+
    \frac{n}{n+1}q_dP\left(J_{d-1}=k+\frac{1}{n}\right)+
    \frac{1}{n+1}q_dP\left(J_{d-1}=k-1\right)\;,
    \label{eq:methods_1_dif_step_q}
\end{equation}
where again it is understood that $P(J_{d-1}=k')=0$ when $k'$ lies outside the domain of $J_{d-1}$.

The diffusion step is then followed by the usual maximization step

\begin{equation}
    P\left(J_d=k\right)=\left(P\left(Q_d\le k\right)\right)^b-\left(P\left(Q_d\le k-1\right)\right)^b\;.
    \label{eq:methods_1_max_step_q}
\end{equation}

\section*{Acknowledgments}
This work is supported by the Howard Hughes Medical Institute (HHMI, ref 55008742), MINECO (Spain; BFU2017-85936-P) and ICREA Academia (2016) to R.M.-B. C.M is supported by a FI fellowship from the Secretariat for Universities and Research of the Ministry of Business and Knowledge of the Government of Catalonia and the European Social Fund.

\section*{Code availability}
 The data generating the results and the C codes to reproduce them, as well as the Matlab codes to generate figures, are available at this public \href{https://github.com/Chiara-Mastro/Deep_imagination_decision_trees}{GitHub repository}.

\bibliographystyle{unsrt}

\bibliography{refs}

\end{document}